\documentclass[10pt,twocolumn,letterpaper]{article}

\usepackage{iccv}
\usepackage{times}
\usepackage{epsfig}
\usepackage{graphicx}
\usepackage{amsmath}
\usepackage{amssymb}
\usepackage{booktabs}

\usepackage[pagebackref=true,breaklinks=true,letterpaper=true,colorlinks,bookmarks=false]{hyperref}

\iccvfinalcopy 

\def\iccvPaperID{8738} 

\ificcvfinal\fi
\begin{document}

\title{ToonTalker: Cross-Domain Face Reenactment}

\author{Yuan Gong\textsuperscript{1}\quad Yong Zhang\textsuperscript{2}\thanks{Corresponding authors.}\quad Xiaodong Cun\textsuperscript{2}\quad Fei Yin\textsuperscript{1}\quad \\Yanbo Fan\textsuperscript{2}\quad Xuan Wang\textsuperscript{3}\quad Baoyuan Wu\textsuperscript{4}\quad Yujiu Yang\textsuperscript{1}\footnotemark[1]
\\ \textsuperscript{1}Shenzhen International Graduate School, Tsinghua University
 \textsuperscript{2}Tencent AI Lab \\ \textsuperscript{3}Ant Group  \textsuperscript{4}The School of Data Science, Shenzhen Research Institute of Big Data, \\ The Chinese University of Hong Kong, Shenzhen (CUHK-Shenzhen)\\
}

\newcommand{\yuangong}[1]{{\textcolor{blue}{[yuangong: #1]}}}
\maketitle
\ificcvfinal\thispagestyle{empty}\fi

\newcommand{\xiaodong}[1]{{\color{red}{[xiaodong: #1]}}}

\begin{abstract}
We target cross-domain face reenactment in this paper, \textit{i.e.,} driving a cartoon image with the video of a real person and vice versa. 
Recently, many works have focused on one-shot talking face generation to drive a portrait with a real video, \textit{i.e.,} within-domain reenactment. 
Straightforwardly applying those methods to cross-domain animation will cause inaccurate expression transfer, blur effects, and even apparent artifacts due to the domain shift between cartoon and real faces. 
Only a few works attempt to settle cross-domain face reenactment. 
The most related work AnimeCeleb~\cite{kim_animeceleb_2022}
requires constructing a dataset with pose vector and cartoon image pairs by animating 3D characters, which makes it inapplicable anymore if no paired data is available.  
In this paper, we propose a novel method for cross-domain reenactment without paired data.   
Specifically, we propose a transformer-based framework to 
align the motions from different domains into a common latent space where motion transfer is conducted via latent code addition.
Two domain-specific motion encoders and two learnable motion base memories are used to capture domain properties. 
A source query transformer and a driving one are exploited to project domain-specific motion to the canonical space.  
The edited motion is projected back to the domain of the source with a transformer. 
Moreover, since no paired data is provided, we propose a novel cross-domain training scheme using data from two domains with the designed analogy constraint. 
Besides, we contribute a cartoon dataset in Disney style. 
Extensive evaluations demonstrate the superiority of our method over competing methods.   
\end{abstract}

\section{Introduction}
\label{sec:intro}
Online video conferences and live streaming are spreading rapidly since the population of smartphones and communication techniques. Speakers sometimes present themselves as cartoon characters for privacy and entertainment consideration.
However, conventional cartoon animation depends on a complex pipeline of 3D modeling and retargeting based on a powerful render engine, which always takes a long period~\cite{morph_animate}. 
This limitation makes it difficult to drive a new model or directly drive a single comic picture.

\begin{figure}
    \centering
    \includegraphics[width=\columnwidth]{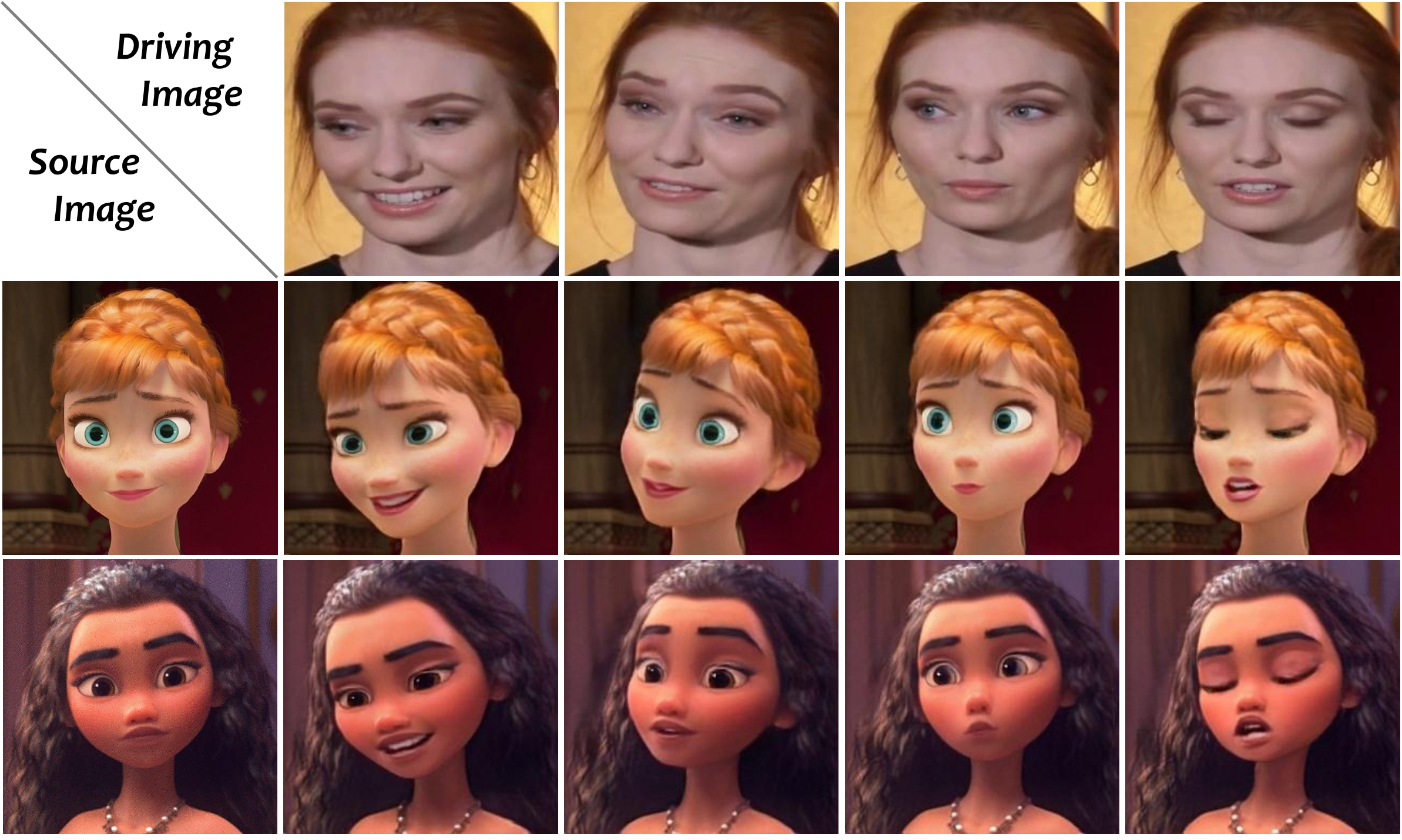}
    \caption{ Cross-domain reenactment examples of our method. Given a cartoon face, we animate it by transferring the pose and expression from a real face. 
    }
    \label{fig:teaser}
\end{figure}

There are many works~\cite{ren_pirenderer_2021, siarohin_first_2020, hong_depth-aware_2022, wang_latent_2022, xing2023codetalker} that focus on one-shot talking face generation, \textit{i.e.,} driving a portrait with a video of a real person, which belongs to within-domain reenactment. 
Their models are trained with a large amount of talking videos and perform well in the same-identity and cross-identity driving tasks. 
However, when those models are  straightforwardly applied to cross-domain reenactment, \textit{i.e.,} driving a cartoon face with a real video as in Fig.~\ref{fig:teaser}, they persistently encounter the following issues, including inaccurate expression transfer, blur effects, and even apparent artifacts around facial components and the background. 
The reason is the distribution shift between different domains. 
There is a large appearance and motion gap between the cartoon and real datasets. 
Though fine-tuning a pre-trained model on a cartoon dataset can improve the visual quality to a certain extent, those issues still cannot be wiped out. 

Only a few methods focus on visually-driven cross-domain face reenactment. 
The early method Recycle-GAN~\cite{bansal2018recycle} overfits two videos for unsupervised video translation, which cannot be used for one-shot reenactment. 
MAA~\cite{xu2022motion} uses keypoints as an intermediate representation and designs an angle consistency loss to mimic the driving motion. 
But only focusing on keypoints will result in ignoring the subtle motion of facial components, especially the mouth and eyes. 
AnimeCeleb~\cite{kim_animeceleb_2022} is based on pre-defined 3D Morphable Models (3DMMs)~\cite{3dmm} and additional pose annotaions\footnote{Note that \texttt{Pose} in~\cite{kim_animeceleb_2022} refers to a 20 dimension~(20D) vector that contains the status of facial expression and head pose.} of cartoon images.
However, it is inapplicable when paired data is not available, which limits its range of use. 
Moreover, the pre-defined pose vector has limited capability of capturing fine-grained facial expressions.

In this paper, we propose a novel method for cross-domain reenactment without the requirement of paired data. 
All we need is a set of real videos and another set of cartoon videos. 
To solve the domain shift problem, we propose a transformer-based framework to align the motions from two domains in a shared canonical latent space. 
Specifically, two motion encoders and two motion bases are designed to discover the domain-specific properties of appearance and motion. 
To align the domain-specific motions in a common latent space, a source query transformer and a target take the responsibility of projecting the motions into a canonical motion space, where  
motion transfer can be conducted via the addition of motion codes. 
The edited motion is then projected from the canonical space back to the domain of the source by a transformer. 
A generator takes the projected motion and the source features as input to synthesize a face in the domain of the source. 

\begin{figure}[t]
  \centering
   \includegraphics[width=0.8\linewidth]{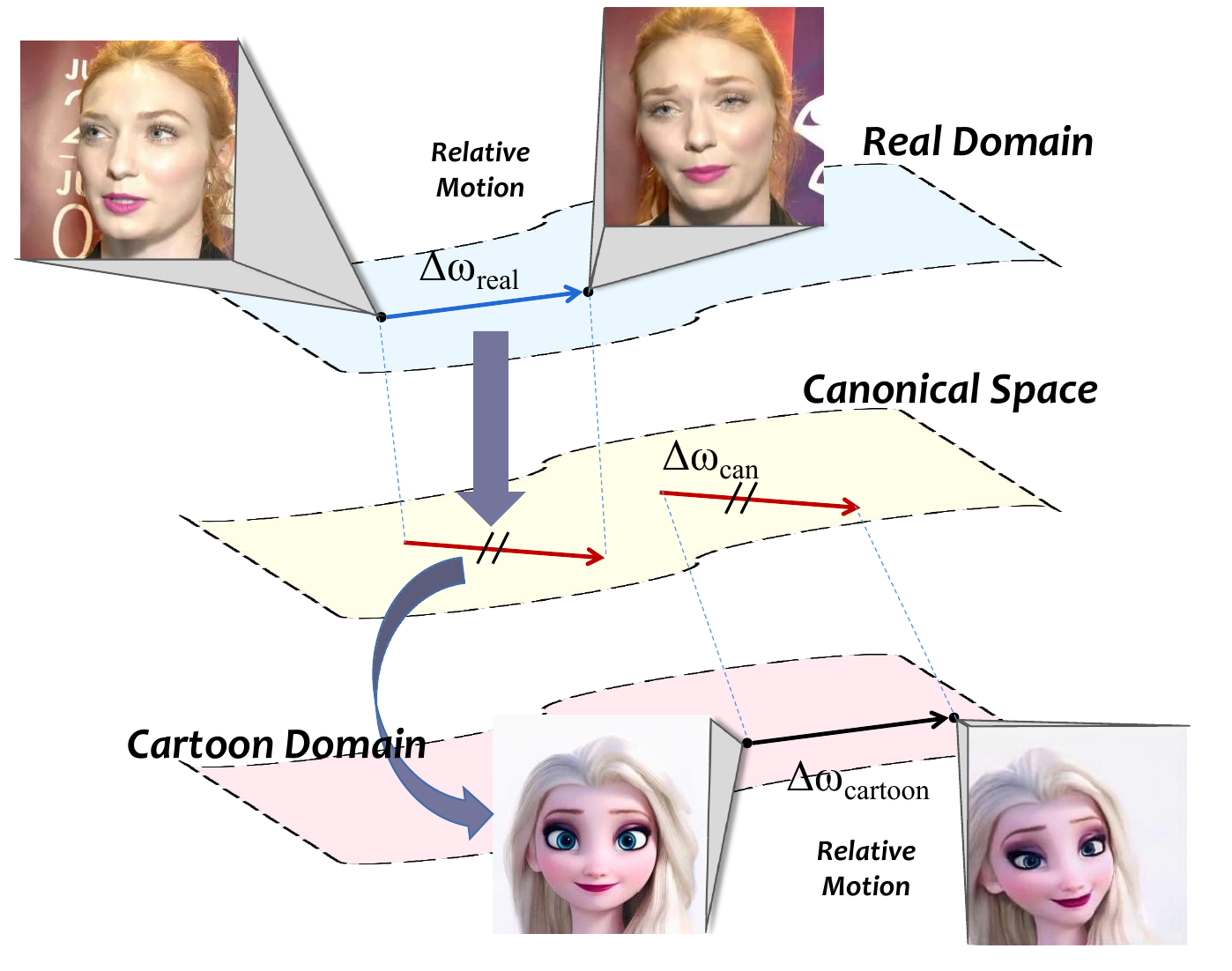}

   \caption{Visualization of analogy constraint. $\Delta\omega_{real}$ and $\Delta\omega_{cartoon}$ represent the relative motion in the real and cartoon domain, respectively. $\Delta\omega_{can}$ represents the projected motion in the canonical space where we can perform cross-domain motion transfer by the addition of motion codes. 
  Given a cartoon face, we utilize the relative motion between two real faces to animate the cartoon one and generate a synthetic cartoon face. 
  We hold that the relative motion between two real faces and that between the cartoon face and the synthetic are equivalent within the canonical motion space. 
   }
   \label{fig:onecol}
\end{figure}

Moreover, to overcome the lack of paired data, we propose a novel cross-domain training scheme. 
The core idea is the analogy constraint (see Fig.~\ref{fig:onecol}). 
Specifically, given two real faces and a cartoon face, we use the relative motion between the two real faces to drive the cartoon one, producing a synthetic cartoon face. 
The relative motion from one real face to the other real one is analogous to that from the cartoon face to the synthetic one. 
We measure the distance between the two types of analogical motions in the aligned canonical motion space. They are supposed to be identical. 

To conduct cross-domain reenactment experiments, we collect a cartoon video dataset in Disney style, which contains 344 videos and about 47k frames in total. 
The cartoon dataset will be released for research purposes. 

Our contributions are as follows:
\begin{itemize}
\item We propose a novel framework for cross-domain face reenactment without using paired data. Several transformers are designed to align the motions of different domains in a common latent space. 
\item We propose a training scheme to compensate for the lack of paired data by using the analogy constraint. 
\item We collect a cartoon video dataset and conduct extensive experiments to demonstrate the superiority of our method in the cross-domain reenactment.
\end{itemize}

\section{Related Work}
\label{sec:related works}

\subsection{Within-domain Face Reenactment}

Most one-shot talking face generation works focus on within-domain face reenactment, \textit{i.e.,} driving a portrait image using another talking head video, where both the driving video and source image is from real humans. 

\textbf{Keypoint-based Methods.}
Many reenactment algorithms~\cite{wu_reenactgan_2018, zakharov_few-shot_2019, zhang_freenet_2020, chen_puppeteergan_2020} are based on pre-defined facial landmarks that are used for motion conversion and repositioning. 
For example, ReenactGAN~\cite{wu_reenactgan_2018} encodes the face into the boundary hidden space defined by the keypoint heat map and transfers motion by a boundary transformer. Although ReenactGAN can produce high-quality target faces, it needs to retrain the model for the new individual images.
FReeNet~\cite{zhang_freenet_2020} trained a keypoint converter to convert the driven expression onto the source face. However, it can only deal with expression change, but not attitude changes, so the generated face image will keep the pose of the source face.

\textbf{3DMM-based Methods.}
A number of works~\cite{ren_pirenderer_2021, doukas_headgan_2021, bounareli_finding_2022, yin_styleheat_2022, zhang2023sadtalker} take advantage of the
3D deformation model~\cite{booth_3d_2016} that can decouple pose, expression, and identity.
For instance, Pirenderer~\cite{ren_pirenderer_2021} mapped the target 3DMM parameters into hidden codes and injected them into the source image features to guide the generation of distorted streams. HeadGAN~\cite{doukas_headgan_2021} renders 3D parameters to images as motion representations. ~\cite{bounareli_finding_2022} map 3DMM parameters to StyleGAN~\cite{karras_style-based_2019, karras_analyzing_2020} latent space, and a transformation matrix is learned to map 3DMM parameters to latent code. StyleHEAT~\cite{yin_styleheat_2022} used 3DMM to generate an optical flow map to change the feature map of StyleGAN. Using 3DMM parameters as motion representation can avoid the problem of video identity information leakage in cross-identity reenactment, but the 3D model only encodes facial regions, such as teeth, eyes, hair and other information that is difficult to synthesize.

\textbf{Unsupervised Methods.}
Another large class of methods~\cite{ferrari_x2face_2018,siarohin_animating_2019, siarohin_first_2020, wang_one-shot_2021, hong_depth-aware_2022, wang_latent_2022, drobyshev_megaportraits_2022} do not use existing face models, but use unsupervised training to obtain facial motion representations. 
A typical method is FOMM~\cite{siarohin_first_2020}, which proposes a first-order motion method by representing the motion as the key points and its local transformation. Face-vid2vid~\cite{wang_one-shot_2021} further extends this method to 3D facial key points. 
DAGAN~\cite{hong_depth-aware_2022} used the self-supervised geometry learning method to obtain the depth map from the video.
Motion transformer~\cite{tao_motion_2022} uses the interactive capabilities of transformer to predict key points. Differing from the above methods, LIA~\cite{wang_latent_2022} does not require explicit structural representation but directly manipulates the latent space of the generative model. Megapotrait~\cite{drobyshev_megaportraits_2022} follows a similar way and adds a super-resolution network for higher resolution.

The goals of these within-domain reenactment methods
are different from ours since they do not need to consider the domain gap. 
They can be trained in a self-supervised way from the same video, while there are no cartoon-real training pairs in our cross-domain settings.

\subsection{Cross-domain Face Reenactment}
Only a few methods focus on cross-domain face reenactment. 
The early study Recycle-GAN~\cite{bansal2018recycle} is based on image translation networks. 
It introduces a recurrent loss by training a temporal predictor to monitor the temporal consistency.
And each character requires two face videos as two domains for training, which cannot be used for one-shot reenactment.
Everythingtalking~\cite{song_everythings_2021} proposes a parametric shape modeling technique parametric models based on Bezier Curve and transfers motion by adapting the human face's control points to the pareidolia face. But it is unable to change the facial posture.
MAA~\cite{xu2022motion} is a method that relies on keypoints and incorporates an angle consistency loss to replicate driving motion. 
However, the differences in keypoints only account for absolute motion transfer, which might significantly alter the appearance of source images and fail to capture intricate facial expressions well.

The most related work AnimeCeleb~\cite{kim_animeceleb_2022} proposes a new cartoon talking-head video dataset. 
It uses a 3D animation model to generate a large number of animated face images and corresponding pose annotations.
To achieve cross-domain reenactment, it proposes to learn the mapping relationship from animation pose vectors to 3DMM parameters by paired data.
One shortcoming of AnimeCeleb is that it becomes inapplicable if no paired data is available, which limits its application scope. 
Moreover, it cannot be generalized to other cartoon styles that require rich expressions. Because AnimeCeleb uses the manually designed 20D vector to represent the status of pose and expression, which is far from enough.  
Differently, our method does not depend on 3DMMs and can be trained without being paired by using a designed cross-domain training scheme.



\begin{figure*}[t]
  \centering
   \includegraphics[width=\linewidth]{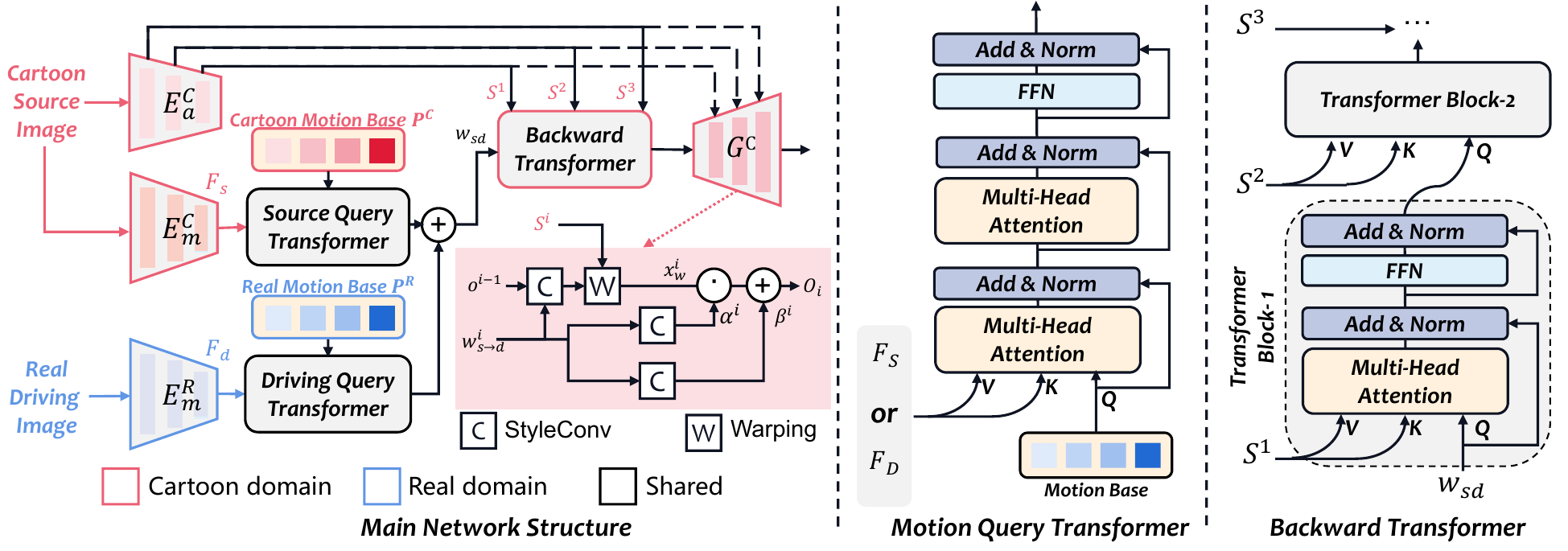}
   \caption{An illustration of our model, which can be mainly divided into three parts: 
   (1) Feature extraction. We use $E_{a}$ and $E_{m}$ to extract local features and latent motion descriptor respectively. 
   (2) Latent space transformation.
   A source query transformer and a driving one are used to project the motion descriptor into a canonical motion space. 
   Motion transfer is conducted via the addition of motion codes in the shared motion space. 
   (3) Image generation. The edited motion is projected back to the domain of the source with the assistant of the source features.  
   A progressive image generator is used to render an image with the projected motion. 
   We also apply a spatial feature transformation (SFT) to the last three high-resolution features to reduce the artifacts caused by feature warping.}
   \label{fig:architecture}
\end{figure*}
\section{Method}
\label{sec:method}
We propose a novel transformer-based framework to align the motions from different domains in a shared canonical motion space for motion transfer. The framework is shown in Fig.~\ref{fig:architecture}. 
The task becomes simple if there exist paired data, \textit{i.e.,} faces in different domains on the same expression and pose. 
Unfortunately, no such dataset is available. 
To settle the lack of paired data, we propose a training scheme with the designed analogy constraint, using faces from two domains jointly.  
We first introduce the architecture of the framework in Sec.~\ref{subsec:architecture}. 
Then, we present the training scheme in Sec.~\ref{subsec:scheme}. 
Finally, we show a two-stage learning strategy to alleviate the issue of imbalanced training data in Sec.~\ref{subsec:finetune}, \textit{i.e.,} the number of real videos is much larger than that of cartoon videos. 

\subsection{Transformer-based Framework}~\label{subsec:architecture}
As shown in Fig.~\ref{fig:architecture}, given a source image and a driving one from two domains, we extract the motion information from the two images with two domain-specific motion encoders. 
Then, a source query transformer and a driving one are used to project the extracted motion information from different domains into the shared canonical motion space, with the domain-specific motion code base and the motion information as input. 
The motion transfer is realized by adding motion codes in the shared motion space. 
A domain-specific backward transformer projects the edited motion code back to the domain of the source image with the assistance of its features extracted via a domain-specific appearance encoder. 
Finally, a generator takes the projected motion code and the source features as input to render an image in the domain of the source. 

Please note that we use the cartoon as the source and the real as the driving for illustration in Fig.~\ref{fig:architecture}. 
We can exchange the roles of the cartoon and the real, \textit{i.e.,} the real as the source while the cartoon is the driving. 
Therefore, in our full framework, we have two appearance encoders, two motion encoders, two backward transformers, two generators, and two motion code bases, which are domain-specific. 
We also have one source query transformer and one driving query transformer, which are shared by two domains. 
The bounding box in color indicates whether a component is domain-specific in Fig.~\ref{fig:architecture}. 
We present the components in detail as follows.

\noindent\textbf{Appearance and Motion Encoders.} 
We have domain-specific encoders to extract appearance and motion information from an image. 
Let $E_a^C$ and  $E_m^C$ denote the appearance and motion encoder for the cartoon domain, respectively, while $E_a^R$ and  $E_m^R$ for the real domain. 
As the roles of the two domains as the source and the target can be exchanged, we use $E_a$ and $E_m$ for simplicity. 
The appearance encoder takes the source image as input and outputs multi-scale spatial features, while the motion encoder extracts motion descriptors from the source and driving image, respectively. 
Thus, the feature extraction can be defined as:
\begin{gather}
  S  = E_a(x_s),  \\
  F_s = E_m(x_s),F_d = E_m(x_d), 
\end{gather} 
where $F_s$ and $F_d$ denote the motion descriptors of the source $x_s$ and the driving $x_d$.  $S=\{S^i\}_{i=1}^N$ represents the appearance features of the source. 

\noindent\textbf{Motion Query Transformer.} 
In addition to the domain-specific motion encoder, we use a motion base for each domain for further discovering domain-specific motion properties. 
The motion base consists of learnable embeddings to memorize typical motions,  
denoted as $p =\{p^k\}^K_{k=1}$, where $p_k\in R^d$ represents the $k$-th motion embedding and $d$ is the dimension. 
We adopt transformer blocks to project the latent descriptor into a canonical motion space due to their capability of capturing long-range dependencies. 
As shown in the middle of Fig.~\ref{fig:architecture}, the embedding is linearly projected to the query features while the latent descriptor from the image is projected to the key and value features. 
Inspired by~\cite{siarohin_first_2020}, since it is challenging to model motion transformation $x_s\rightarrow x_d$ straightforwardly, we choose to model transformation $x_s\rightarrow x_r$ and $x_r\rightarrow x_d$ separately by leveraging a virtual reference $x_r$:
\begin{align}
    w_{s\rightarrow d} = w_{s\rightarrow r} + w_{r\rightarrow d}, 
\end{align}
where $w_{s\rightarrow r} = T_{s\rightarrow r}(F_s, p) $ and $w_{r\rightarrow d} = T_{r\rightarrow d}(F_d, p)$. $w_{s\rightarrow r}$,$w_{r\rightarrow d}$, and $w_{s\rightarrow d}$ are three motion codes in canonical motion space. 
$T_{s\rightarrow r}$ denotes the source query transformer that models the motion transformation\textit{ from the source to the reference}. 
While $T_{r\rightarrow d}$ denotes the driving query transformer that models the motion transformation \textit{from  the reference to the driving}. 
The two query transformers are shared by the two domains. 

\noindent\textbf{Backward Transformer.} 
As we obtain the motion transformation $w_{s\rightarrow d}$ in the canonical motion space, the next step is to project the motion code back to the domain of the source image. 
We use a domain-specific backward transformer to take over this task with the assistance of the information of the source image, \textit{i.e.,} the multi-scale features $S$ of the source image from the appearance encoder $E_a$. 
We have two such transformers, \textit{i.e.,} $T_B^C$ and $T_B^R$, and take the notation $T_B$ for simplicity.
To let the motion code obtain information from image features in different resolutions, we design the backward transformer as a stack of three transformer blocks. 
As shown on the right of Fig.~\ref{fig:architecture}, in each transformer block, the query features are derived from the output of the previous module, while the key and value features are computed from the corresponding multi-scale features. 
The first query features are the linear projection of the motion transformation $w_{s\rightarrow d}$ from the canonical space. 
And the resolutions of the feature maps are 16$\times$16, 32$\times$32, and 64$\times$64, respectively. 

\noindent\textbf{Image Generator.} 
After obtaining the projected motion code in the source domain, we use a domain-specific generator to progressively warp and refine the source feature maps layer by layer conditioned on the motion code $w_{s\rightarrow d}$. 
We have two generators, \textit{i.e.,} $G^C$ and $G^R$, and take the notation $G$ for simplicity.
Considering that warping them only will make the final output suffer from artifacts, we add refinement modules in the last three layers of the generator to enhance three high-resolution feature maps. 
Given the motion code ${w_{s\rightarrow d}}$ and the previous feature maps $o^{i-1}$, a StyleConv block is used to predict a flow field that is used to warp the feature map $S^i$. 
With the warped feature map $f_w^i$ and the motion code ${w_{s\rightarrow d}}$ as input, the spatial feature transformation (SFT) for refinement is defined as:
\begin{align}
    o^i = \alpha^i \odot f_w^i + \beta^i, 
\end{align}
where $\alpha^i = \text{StyleConv}(f_w^i, {w_{s\rightarrow d}}) $ is the spatial scale and $\beta^i = \text{StyleConv}(f^i_w, {w_{s\rightarrow d}})$ is the bias.

\subsection{Cross-domain Training Scheme}~\label{subsec:scheme}
If there are paired data where cartoon and real video pairs share the same head pose and expression, cross-domain training can be settled simply following the way of training a conventional within-domain model. 
However, no such paired data is available. 
To tackle this issue, we propose an analogy constraint to train the model with unpaired cartoons and real videos. 

\noindent\textbf{Analogy Constraint.} 
As shown in Fig.~\ref{fig:onecol}, the core idea is to project the relative motions of different domains into the canonical motion space. 
For instance, given a tuple consisting of two real faces and a cartoon one, we can calculate the relative motion from one real face to the other real one. 
Then, the relative motion can be used to animate the cartoon face to generate a synthetic cartoon face. 
We can compute the relative motion from the cartoon face to the synthetic cartoon one. 
The two types of relative motions are supposed to be identical in canonical motion space. 

In the canonical motion space, following~\cite{siarohin_first_2020}, the relative motion transfer can be defined as: 
\begin{equation}
	w_{s\rightarrow d} = (w_{s\rightarrow r} + w_{r\rightarrow s}) + (w_{r\rightarrow d} -  w_{r\rightarrow 1}), 
\end{equation}
where `1' refers to a selected frame in the driving video. 
For illustration, let $x_a^R$ and $x_b^R $ denote two real images and $x_s^C$ denote  a cartoon image. 
Then, the relative motion transfer is applied to the cartoon, which can be defined as: 
\begin{equation}
	w_{s\rightarrow g} = (w_{s \rightarrow r} + w_{r\rightarrow s}) + (w_{r\rightarrow b} -  w_{r\rightarrow a}), 
\end{equation}
where $w_{s\rightarrow g}$ represents the motion transformation from $x_s^C$ to $x_g^C$. 
`$g$' indicates the synthetic cartoon face. 
Each item can be defined as follows:
\begin{equation}
	\begin{split}
	w_{s\rightarrow r} &= T_{s\rightarrow r}(E_m^C(x_s^C), p^C), \\
        w_{r\rightarrow s} &= T_{r\rightarrow d}(E_m^C(x_s^C), p^C), \\
        w_{r\rightarrow {a}} &= T_{r\rightarrow d}(E_m^R(x_{a}^R), p^R), \\
        w_{r\rightarrow {b}} &= T_{r\rightarrow d}(E_m^R(x_{b}^R), p^R).  
	\end{split}
\end{equation}
With the motion code $w_{s\rightarrow g}$ as well as the source features $S$, the generator $G^C$ can produce a synthetic cartoon face $x_g^C$, \textit{i.e.,}
$x_g^C = G^C(T_B^C(w_{s\rightarrow g}), E_a^C(x_s^C))$. 
Then, we can calculate the motion code of the synthetic cartoon face in the canonical motion space, \textit{i.e.,}
\begin{equation}
    w_{r\rightarrow g} = T_{r\rightarrow d}(E_m^C(x_g^C), p^C). 
\end{equation}
The relative motion between the cartoon face and the synthetic one is $ ( w_{r\rightarrow g} -  w_{r\rightarrow s})$, which is supposed to be identical to the relative motion between the two real faces, \textit{i.e.,} $( w_{r\rightarrow b} -  w_{r\rightarrow a})$. 
Therefore, we can measure the distance between the two types of relative motions by 
\begin{equation}
    \mathcal{L}_{motion}^{R\rightarrow C} = \mathbb{E}[\left\| (w_{r\rightarrow g} - w_{r\rightarrow s}) - (w_{r\rightarrow b} -  w_{r\rightarrow a}) \right\|_1]. 
\end{equation}
Similarly, given two cartoon faces and one real face, we have $\mathcal{L}_{motion}^{C\rightarrow R} $ following the same formulation. 
Further, to maintain the high fidelity of face generation, we impose adversarial loss.
The adversarial loss can be defined as:
\begin{equation}
	\begin{split}
	\mathcal{L}_{adv}^R &= \mathbb{E}_{x_g^R}[- \text{log}(D^R(x_g^R))], \\
        \mathcal{L}_{adv}^C &= \mathbb{E}_{x_g^C}[-\text{log}(D^C(x_g^C))],
	\end{split}
\end{equation}
where $D^R$ and $D^C$ are two discriminators for the real and cartoon domains, respectively.

\noindent\textbf{Reconstruction.} 
Besides the analogy constraint that emphasizes the interaction between two domains, we can also perform within-domain reenactment to assist the model learning. Specifically, we can sample two images within the same domain as the source and the driving, \textit{e.g.,} a pair of $x_s^R$ and $x_d^R$ or a pair of $x_s^C$ and $x_d^C$. 
The reconstruction loss of the real pairs can be used to update the parameters of the real domain-related components as well as the shared motion query transformers. 
Similarly, the reconstruction loss of the cartoon pairs is for the corresponding components.  
We exploit the $L_1$ loss $\mathcal{L}_1(x_{s\rightarrow d}, x_d) $, the perceptual loss $\mathcal{L}_{per}(x_{s\rightarrow d}, x_d) $, and the GAN loss $\mathcal{L}_{adv}(x_{s\rightarrow d}, x_d)$. 
Hence, the objective of reconstruction can be defined as : 
\begin{equation}
    \begin{split}
        \mathcal{L}_{rec} =~~& \mathcal{L}_1  
        + \lambda_{per} \mathcal{L}_{per} 
        + \mathcal{L}_{adv}, 
    \end{split}
\end{equation}
where we omit the domain notation R\&C for simplicity.

The overall objective is defined as follows:
\begin{equation} \label{eq:influence}
    \mathcal{L} = \mathcal{L}_{rec}^R + \mathcal{L}_{rec}^C + \lambda_m * (\mathcal{L}_{motion}^{R\rightarrow C} + \mathcal{L}_{motion}^{C\rightarrow R}) + \mathcal{L}_{adv}^C + \mathcal{L}_{adv}^R,
\end{equation}
where $\lambda_m $ is the trade-off hyperparameter.

\begin{figure*}[t]
\centering
\includegraphics[width=\linewidth]{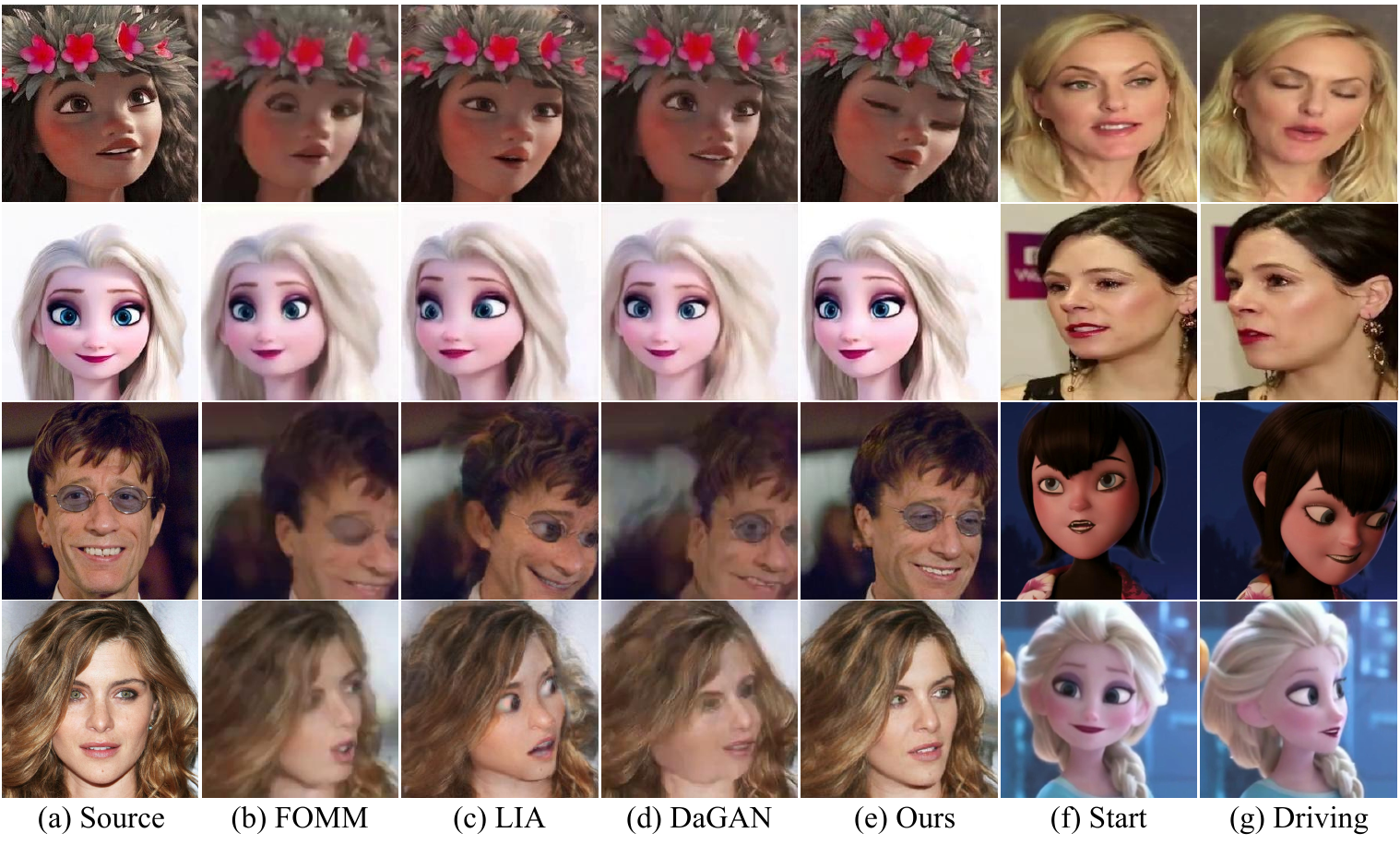}
\caption{Qualitative comparisons with \textit{state-of-the-art} methods on cross-domain reenactment. 
Competing methods are finetuned with the cartoon dataset. Hence all models are trained on the same data. It can be seen that our results have superior image quality and motion consistency. More visual results are in the supplementary.}
\label{fig:cross-domain}
\end{figure*}

\subsection{Two-stage Learning on Imbalanced Data.}~\label{subsec:finetune}
The number of real videos is much larger than that of cartoon videos. 
Directly training with mixed imbalanced data cannot promote the performance of cross-domain reenactment because real faces dominate the data distribution. 
To alleviate the issue of the imbalanced dataset, we propose to learn the model in two stages. 

In the first stage, we train the model with real videos by performing the reconstruction task, through which the model can learn the general knowledge of face motion from a large number of talking head videos.The obtained model cannot accurately transfer motion from real faces to cartoon ones due to domain shifts, especially facial expressions. 

In the second stage, we copy the domain-specific components of the real domain as initialization for those of the cartoon domain. 
Then, we jointly learn all the components in the framework with the analogy constraint and the reconstruction loss.

Please note that our final model can be used to reenact a cartoon with a real video as well as to reenact a real face with a cartoon video.


\section{Experiments}
\label{sec:experiment}

\subsection{Settings}
\noindent \textbf{Datasets.} 
For cross-domain reenactment, we collect a cartoon video dataset in Disney style which contains 344 videos with an FPS of 30 and the total number of frames is about 47k.
To train our model, we take a subset of videos from the VoxCeleb~\cite{nagrani_voxceleb_2017} dataset containing $17$K videos, and a subset from self-collected cartoon videos containing $289$ videos. For evaluation, we use the rest 473 videos from the VoxCeleb~\cite{nagrani_voxceleb_2017} dataset and 55 videos from the cartoon dataset.
We also use a set of images from CelebA-HQ~\cite{lee2020maskgan} for cross-identity reenactment evaluation.  

\vspace{2pt}
\noindent \textbf{Metrics.}
We evaluate the performance from four aspects, \textit{i.e.,} \textit{reconstruction}, \textit{fidelity}, \textit{identity preservation}, and \textit{motion transfer}.
\textbf{$L_1$},  Learned Perceptual Image Patch Similarity~(\textbf{LPIPS})\cite{zhang2018unreasonable}, and Peak Signal-to-Noise Ratio~(\textbf{PSNR}) are used to measure the reconstruction accuracy. 
Frechet Inception Distance~(\textbf{FID})~\cite{heusel2017gans} and Cumulative probability of Blur Detection~(\textbf{CPBD})~\cite{narvekar2011no} are used for measuring generation realism. 
The cosine similarity~(\textbf{CSIM}) between synthetic and source images through ArcFace~\cite{deng2019arcface} is used for identity preservation. 
To evaluate the accuracy of motion transfer, following~\cite{hong_depth-aware_2022}, we use Average Keypoint Distance~(\textbf{AKD}) and Average Expression Distance~(\textbf{AED}) to compute the distance between synthetic and target images in terms of expression and pose. 

\vspace{2pt}
\noindent \textbf{Competing methods.} 
We compare our method with several recent state-of-the-art methods, including FOMM~\cite{siarohin_first_2020}, DAGAN~\cite{hong_depth-aware_2022}, and LIA~\cite{wang_latent_2022}. 
FOMM is a typical motion transfer method based on unsupervised keypoints. 
DAGAN introduces depth into the estimation of key points and has a similar formulation of motion transfer as FOMM. 
LIA is a representative method that uses a reference space similar to FOMM but conducts motion transfer in a latent space.
For a fair comparison, we initialize the models using their pretrained official models and finetune them on the cartoon dataset. Therefore, we learn with the same data.
Furthermore, we also compare with MAA~\cite{xu2022motion}, a keypoint-based cross-domain motion transfer method. 


\vspace{2pt}
\noindent \textbf{Implementation Details.} 
The model is trained in two stages to handle the imbalanced dataset where real videos are much more than cartoons. 
In the first stage, we trained the model on the training set of VoxCeleb for $450$K iterations,
where the batch size is $64$ and the optimizer is Adam with an initial learning rate of $5\times 10^{-4}$. 
In the second stage, we fine-tune the model with Disney videos with $10$K iterations, 
where the batch size is $8$ and the optimizer is Adam with an initial learning rate of $1\times 10^{-4}$. 
The hyper-parameters are set as $\lambda_{per} = 10$ and $\lambda_m = 20$ to balance generative fidelity and motion consistency.
A detailed description of model components is in the supplementary.

\begin{figure}
    \centering
    \includegraphics[width=0.95\columnwidth]{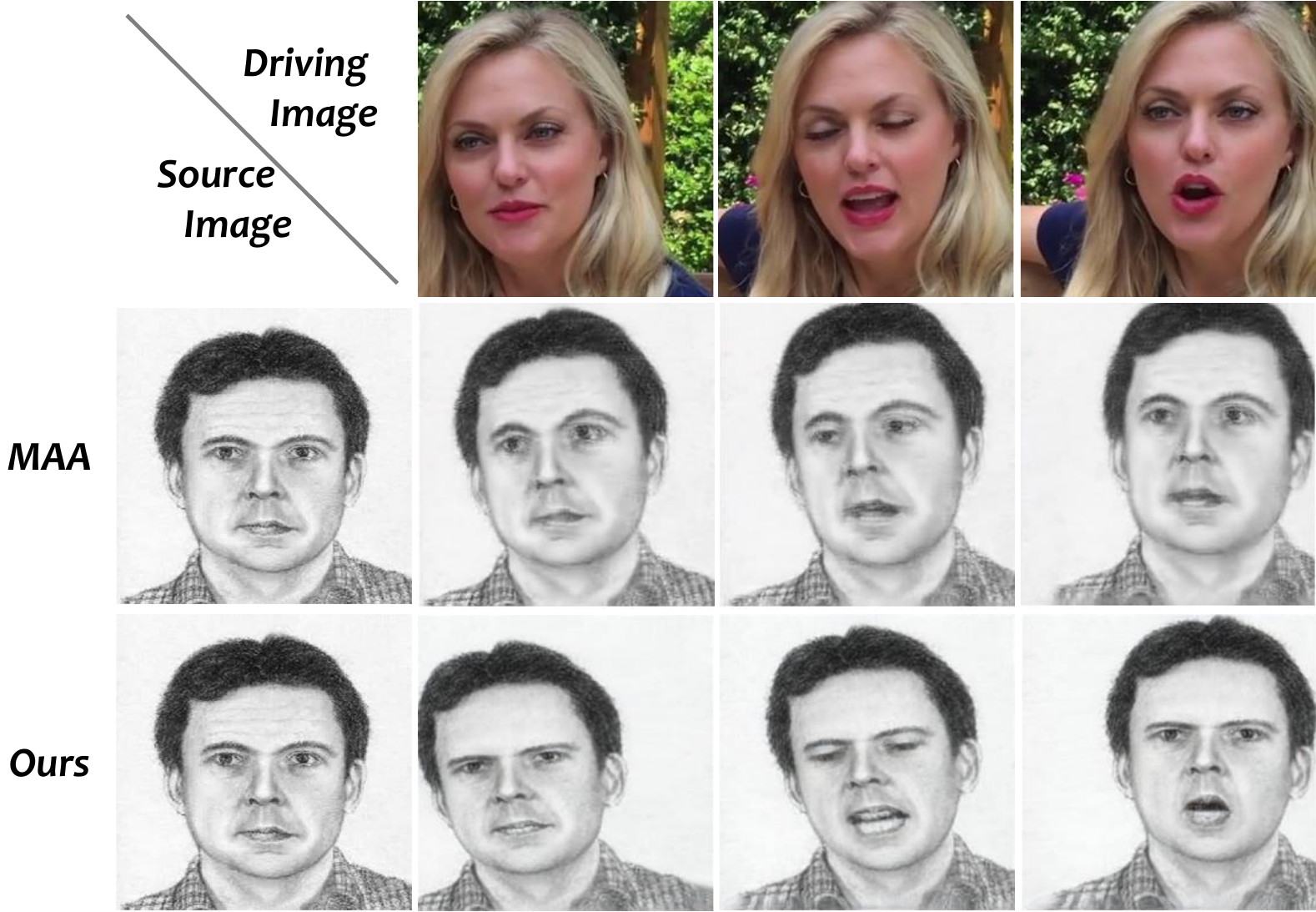}
    \caption{Qualitative comparisons with MAA~\cite{xu2022motion}. Our method can better preserve the face shape of the original image. And more accurately imitated the driving image, especially the eyes and mouth.
    }
    \label{fig:MAA}
\end{figure}

\begin{table}[]
\begin{center}
\setlength{\tabcolsep}{1.5mm}{
\begin{tabular}{cccccc}
\toprule
      & \multicolumn{2}{c}{Real$\rightarrow{}$Cartoon} & \multicolumn{3}{c}{Cartoon$\rightarrow{}$Real} \\
      \cline{2-6}
      & FID$\downarrow{}$      & CPBD$\uparrow{}$      & FID$\downarrow{}$      &CSIM$\uparrow{}$        & CPBD$\uparrow{}$  \\\hline
FOMM  & 30.369                 & 0.0436                & 51.929                 &0.536                   & 0.0626                \\
LIA   & 30.442                 & 0.0493                & 48.611                 &0.515                   & 0.0727                \\
DaGAN & 29.779                 & 0.0735                & 47.222                 &0.561                       & 0.0985                \\
Ours  & \textbf{28.531}        & \textbf{0.0840}       & \textbf{31.852}        &\textbf{0.655}                   & \textbf{0.1230}       \\ \bottomrule
\end{tabular}}
\end{center}
\caption{Quantitative comparison on cross-domain reenactment.
}
\label{tab:cross-domain}
\end{table}

\subsection{Cross-Domain Face Reenactment}
As mentioned in Sec.~\ref{subsec:finetune}, our method can handle the bidirectional cross-domain reenactment. 
We execute motion transfer task for evaluation \textit{i.e.},  \texttt{real$\to$cartoon} and \texttt{cartoon$\to$real}.
The former represents driving a cartoon face with a real video, while the latter represents driving a real face with a cartoon video. 
As for \texttt{real$\to$cartoon}, 60 videos from VoxCeleb are used to drive 300 cartoon images from the cartoon dataset.
For a fair comparison, all competing methods are fine-tuned on the cartoon training set. 
As for \texttt{cartoon$\to$real}, 55 cartoon videos are employed to drive 275 images of different identities from CelebA-HQ~\cite{lee2020maskgan}.

\vspace{2pt}
\noindent \textbf{Qualitative Evaluation.} 
The results of \texttt{real$\to$cartoon} are shown in the first two rows of Fig.~\ref{fig:cross-domain}.
First, our method achieves better image quality than other methods in terms of sharpness and artifacts. 
Second, our method outperforms others in motion transfer, especially facial expression. 
For instance, in the first row, the eyes of the cartoon should be closed according to the driving image.
Only our method mimics driving while the eyes of others stay open.  

The results of \texttt{cartoon$\to$real} are shown in the last two rows of Fig.~\ref{fig:cross-domain}.
We can observe a similar phenomenon as the \texttt{real$\to$cartoon} case in terms of image quality, where our method works the best. 
FOMM and DaGAN become underperformed when there is a large head pose difference between the source and driving, as shown in the third and fourth rows. 
LIA is better than them but still worse than ours. 
For motion and identity preservation, our method surpasses others. 

The comparison with the cross-domain motion transfer method MAA~\cite{xu2022motion} is shown in Fig.~\ref{fig:MAA}. As the source code of MAA is not released, its results are adapted from the original paper.  
Our method has better, especially in the mouth, eyes and image quality. 
For instance, in the third column, MAA ignores the motion of the eyes, whereas our method accurately mimics the driving image. 
In the last column, our mouth shape is more consistent with the driving one.


\vspace{2pt}
\noindent \textbf{Quantitative Evaluation.} 
The quantitative results are reported in Tab.~\ref{tab:cross-domain}. 
Our method achieves the best FID in the two tasks, indicating that the synthetic results are most consistent with the source distribution. 
Our method also performs the best in sharpness and identity preservation, \textit{i.e.,} the highest CPBD and CSIM,  which is consistent with the observation in visual results. 
These results demonstrate that our model can capture the appearance and motion properties of different domains and the motion transfer in the canonical motion space works well. 

\vspace{2pt}
\noindent \textbf{User Study.}
To further evaluate the visual quality, we conduct a user study by asking 20 human raters to answer 20 multiple-choice questions. 
In each question, a rater chooses the best from 4 synthetic cartoon videos generated by the three competing methods and ours. 
Our method is the most favorable with a selection rate of $72.75\%$.
While the rates are $13.75\%$, $9\%$, and $4.5\%$ for LIA, DaGAN, and FOMM.

\begin{figure}[t]
\centering
\includegraphics[width=\linewidth]{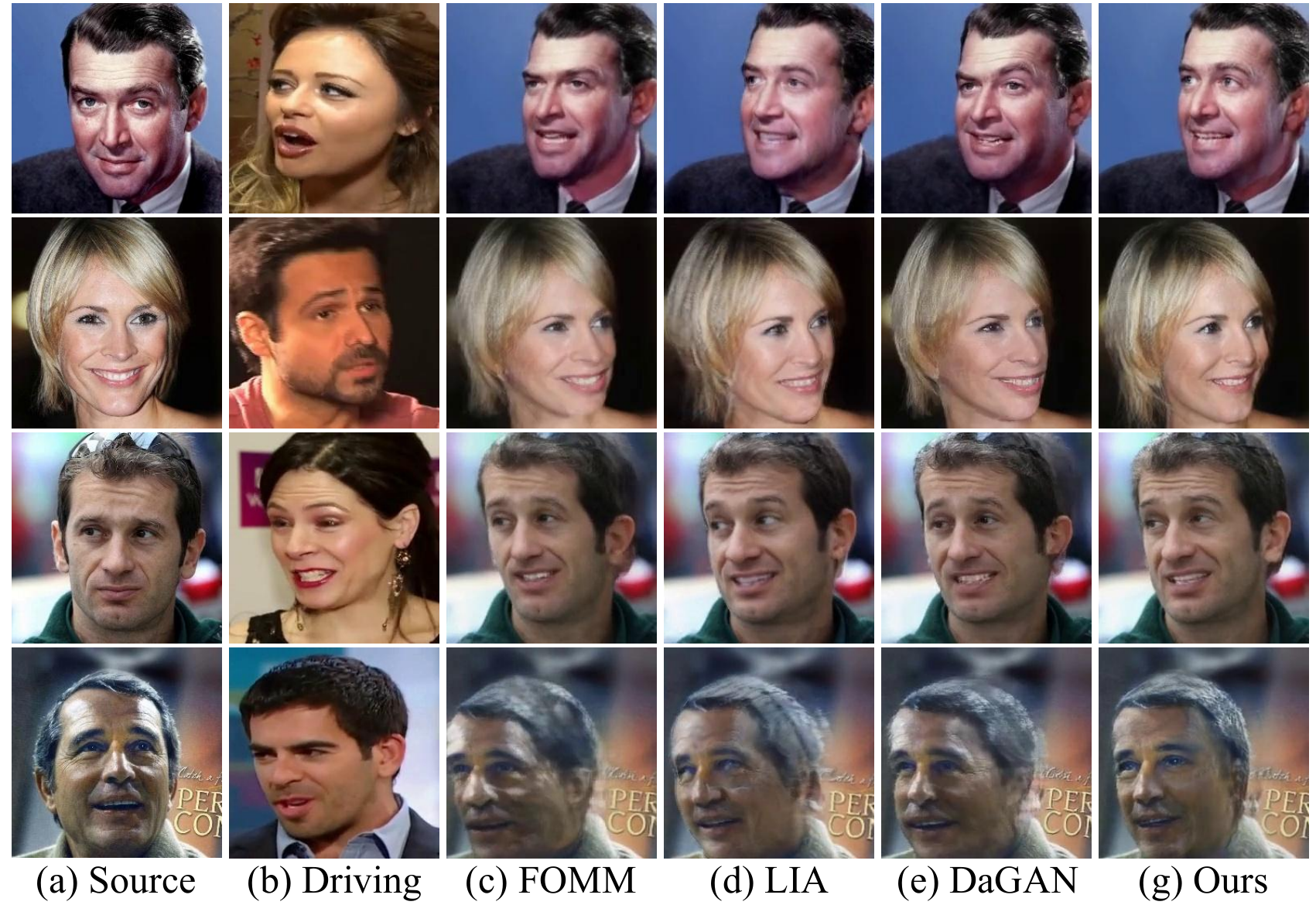}
\caption{Qualitative comparisons on within-domain reenactment. Our method achieves better image quality than other methods in terms of sharpness, distortion and artifacts.}
\label{fig:cross-id}
\end{figure}

\begin{table}[]
\begin{center}
\setlength{\tabcolsep}{1.2mm}{
\begin{tabular}{llllll}
\toprule
           & L1$\downarrow{}$ & LPIPS$\downarrow{}$ & CPBD$\uparrow{}$  & AKD$\downarrow{}$  & AED$\downarrow{}$   \\ \hline
FOMM       & 0.0466           & 0.139               & 0.0308            & 1.388              & 0.142          \\
LIA        & 0.0461           & 0.123               & 0.0372            & 1.512              & 0.153          \\
DaGAN      & \textbf{0.0442}  & \underline{0.122}      & \underline{0.0415}   & \textbf{1.300}     & \textbf{0.129} \\
Ours       & \underline{0.0457}  & \textbf{0.121}      & \textbf{0.0429}   & \underline{1.335}     & \underline{0.139}          \\ \bottomrule
\end{tabular}}
\end{center}
\caption{Same-identity reconstruction in real domain}
\label{tab:same-id}
\end{table}

\begin{table}[]
\begin{center}
\setlength{\tabcolsep}{4.4mm}{
\begin{tabular}{lllll}
\toprule
      & FID$\downarrow{}$    & CSIM$\uparrow{}$    & CPBD $\uparrow{}$  \\ \hline
FOMM  & 54.411               & \underline{0.541}   & 0.0689 \\
LIA   & 53.161               & 0.522               & 0.0787 \\
DaGAN & \underline{49.920}   & \textbf{0.548}      & \underline{0.1069}  \\
Ours  & \textbf{48.713}      & 0.526               & \textbf{0.1132}  \\ \bottomrule
\end{tabular}}
\end{center}
\caption{Cross-identity reenactment in real domain}
\label{tab:corss-id}
\end{table}
\begin{table}[]
\begin{center}
\setlength{\tabcolsep}{2.7mm}{
\begin{tabular}{lllll}
\toprule
                    & FOMM        & LIA         & DAGAN         & Ours   \\ \hline
FID$\downarrow{}$   & 40.853      & 34.066      & 36.218        & \textbf{27.467} \\
CPBD$\uparrow{}$    & 0.0368      & 0.0523      & 0.0598        & \textbf{0.0846} \\ \bottomrule
\end{tabular}}
\end{center}
\caption{Cross-identity reenactment in cartoon domain.}
\label{tab:cartoon}
\end{table}

\subsection{Within-Domain Face Reenactment}
Besides the cross-domain evaluation, we also compare our method with competing methods in the scenario of within-domain reenactment though our method is designed for the cross-domain case. 
For the same-identity evaluation, we use 200 test videos from VoxCeleb. 
For cross-identity evaluation, we use the 200 videos to drive $1,000$ images from CelebA-HQ.
Each video drives five images. 
The obtained $1,000$ videos are used to compute the metrics. 
Note that we use the released pre-trained models of competing methods here. 

The quantitative results are shown in Tab.~\ref{tab:same-id} and Tab.~\ref{tab:corss-id}. 
For same-identity evaluation, our method is overall comparable with DaGAN and better than LIA and FOMM. 
For cross-identity evaluation, our method achieves the best FID and CPBD. Our CSIM is comparable with others. 

Since the visual difference among different methods in same-identity reenactment is marginal, we only present a visual comparison of cross-identity reenactment in Fig.~\ref{fig:cross-id}. 
It can be observed that our results have the best image quality, presenting more details in the face area, \textit{e.g.,} eyes and teeth. 
Besides, our method is comparable with LIA  in preserving face shape and sometimes even better (see the last row in Fig.~\ref{fig:cross-id}).  

In addition, we also perform a cross-identity face reenactment in the cartoon domain.  
We use 55 cartoon videos to drive 275 cartoon images.
Results are shown in Tab.~\ref{tab:cartoon}. 
Our method outperforms others in both FID and CPBD.

\begin{figure}[t]
\centering
\includegraphics[width=\linewidth]{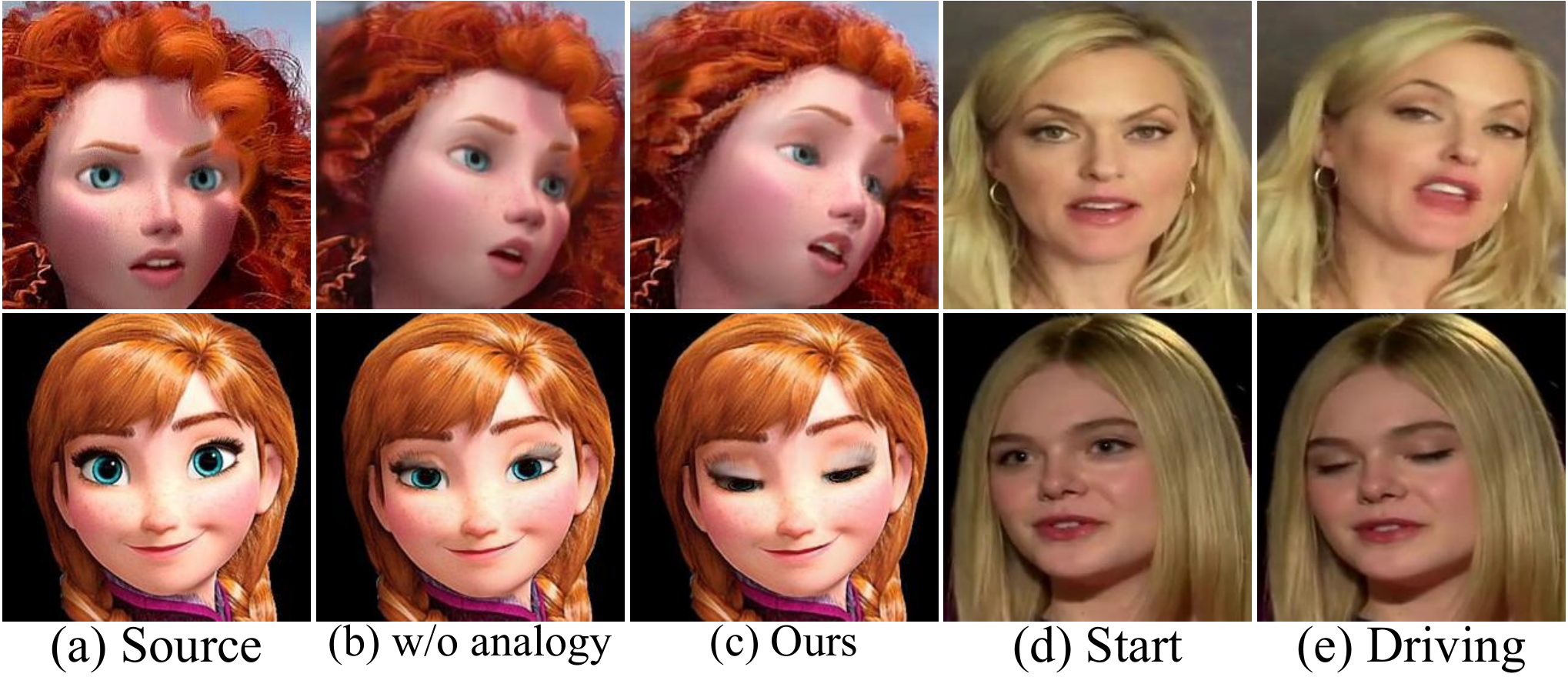}
\caption{Qualitative ablation study on the analogy constraint. We observe that the image quality and motion consistency is much better by using analogy constraint.}
\label{fig:ablation_analogy}
\end{figure}

\subsection{Ablation Study}

To evaluate the training scheme, we perform an ablation study by dropping the analogy constraint. 
The quantitative results are shown in Tab.~\ref{tab:finetune}. 
`\texttt{w/o analogy}' represents dropping the analogy constraint, \textit{i.e.,} finetuning the obtained model from the first stage with the cartoon dataset using only $\mathcal{L}_{rec}^R $ and $ \mathcal{L}_{rec}^C$ in Eq.\ref{eq:influence}.
It can be observed that dropping the analogy constraint causes overall performance degradation, especially for the \texttt{cartoon$\to$real} task. 
The qualitative evaluation is shown in Fig.~\ref{fig:ablation_analogy}. 
In the second row, the result without analogy constraints cannot accurately mimic the motion of the driving image.
The eyes of the cartoon should be closed, but it is opened when dropping constraints.
The dropping operation has a side effect on the motion consistency, which indicates that the analogy constraint contributes. 

To further analyze the influence of analogy constraint, we perform an ablation on the trade-off hyperparamter $\lambda_m$. 
As is shown in Tab.~\ref{tab:codeloss},
it can be observed that $\lambda_m$ trades off the performance of the two cross-domain tasks, \textit{i.e.,} \texttt{Cartoon$\to$Real} and \texttt{Real$\to$Cartoon}.
We empirically choose $\lambda_m=20$ in practical.

We conduct ablation studies to demonstrate the effectiveness of several modules in the architecture, \textit{i.e.,} the query transformer, the backward transformer, and the spatial feature transformation. 
The ablation results are shown in Tab.~\ref{tab:ablation_structure}. The metric is FID. 
``mlp" indicates that we replace the query transformer with multi-layer perception (MLP). ``w/o back" denotes that we drop the backward transformer. And ``w/o sft" drops the spatial feature transformation module. 
In Tab.~\ref{tab:ablation_structure}, we can observe that query transformer brings significant improvement in the  \texttt{Real$\to$Cartoon} case. 
It verifies that the proposed query transformer can project the latent descriptor from different domains into a canonical motion space. 
Additionally, removing the backward transformer and sft will  damage the quality of generated images. 
%


\begin{table}[]
\begin{center}
\setlength{\tabcolsep}{2.9mm}{
\scalebox{0.94}{
\begin{tabular}{lllll}
\toprule
       & \multicolumn{2}{c}{Real $\rightarrow$Cartoon}  & \multicolumn{2}{c}{Cartoon $\rightarrow$ Real}\\
       \cline{2-5}
       & FID$\downarrow{}$                   & CPBD$\uparrow{}$   & FID$\downarrow{}$   & CPBD$\uparrow{}$      \\ \hline
our full model     & \textbf{28.531}         & \textbf{0.0840}     & \textbf{31.852}     & \textbf{0.123}             \\
w/o analogy        & 29.230             & 0.0746                  & 43.353              & 0.107            \\
\bottomrule
\end{tabular}}}
\end{center}
\caption{Ablation study on the analogy constraint.}
\label{tab:finetune}
\end{table}
\begin{table}[]
\begin{center}
\setlength{\tabcolsep}{2.5mm}{
\begin{tabular}{lllll}
\toprule
                             & 0             & 20            & 40            & 100 \\ \hline
Real$\rightarrow{}$Cartoon   & 26.885        & 28.531        & 32.094        & 41.936   \\
Cartoon$\rightarrow{}$Real   & 33.463        & 31.852        & 30.176        & 28.166   \\ \bottomrule
\end{tabular}}
\end{center}
\caption{Ablation study on hyper-parameter $\lambda_m$.}
\label{tab:codeloss}
\end{table}
\begin{table}[]
\begin{center}
\begin{tabular}{lllll}
\toprule
                           & Full       & mlp          & w/o back   & w/o sft  \\ \hline
Real$\rightarrow{}$Cartoon & 28.531     & 116.740      & 28.904     & 31.291 \\
Cartoon$\rightarrow{}$Real & 31.852     & 31.885       & 35.265     & 41.077 \\ \bottomrule
\end{tabular}
\end{center}
\caption{Ablation study on model architecture.}
\label{tab:ablation_structure}
\end{table}

\begin{figure}[ht]
    \centering
    \includegraphics[width=0.6\columnwidth]{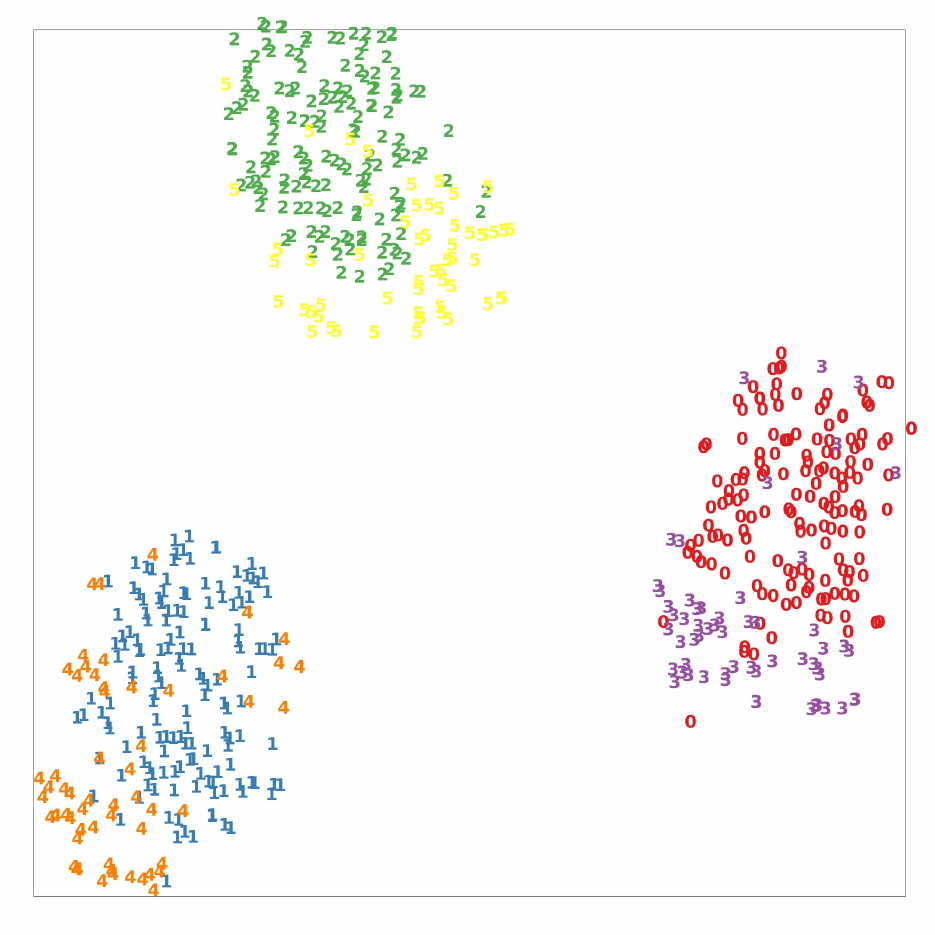}
	\caption{Visualizing the canonical motion space. We visualize the t-SNE plot of motion codes in canonical space. $<$0, 1, 2$>$ and $<$3, 4, 5$>$ are from the relative motions of real domain and cartoon domain, respectively. And $<$0,3$>$, $<$1,4$>$, and $<$2,5$>$ are driven by the same template, respectively.}
	\label{fig:tsne}
\end{figure}

We also conduct a visualization experiment to verify the proposed canonical space. Since no paired data is available, we select three real images with different poses and expressions as templates to drive real and cartoon images, generating 450 real image pairs and 150 cartoon pairs. Each pair contains faces before and after driving. We compute their relative motions in the common motion space. In Fig.~\ref{fig:tsne}, 0, 1, and 2 represent real pairs, while 3, 4, and 5 represent cartoon pairs. $<$0,3$>$, $<$1,4$>$, and $<$2,5$>$ are driven by the same template, respectively. This shows that the motions of the two domains are aligned.

\section{Conclusion}
\label{sec:conclusion}
We propose a novel transformer-based framework for cross-domain face reenactment. 
To settle the domain shift issue, we propose to align the motions of different domains in a canonical motion space. 
A set of domain-specific and domain-shared components are designed for motion alignment. 
Motion transfer can be implemented by adding motion codes in the shared motion space. 
Since there is no paired data for training, we propose a cross-domain training scheme with the analogy constraint using the two domain datasets. 
To alleviate the influence of imbalanced data, we use a two-stage learning strategy. 
\begin{figure}[]
\centering
\includegraphics[width=\columnwidth]{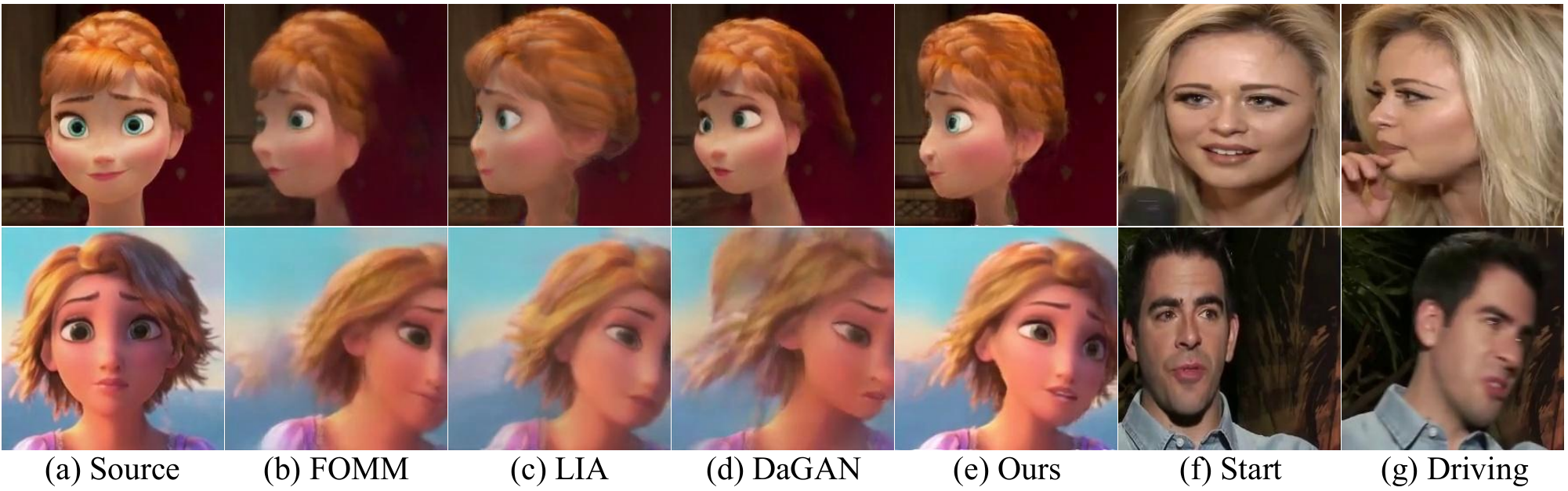}
\caption{Failure cases. We observed that it is challenging to handle extreme poses.}
\label{fig:failure}
\end{figure}
Extensive experiments are conducted to demonstrate the superiority of our method over competing methods in cross-domain reenactment.    
\\

\noindent \textbf{Limitations}. Like other SOTA models, our model cannot handle extreme poses well. As is shown in Fig~\ref{fig:failure}, our method suffers from artifacts and deformations when the driving image is an extreme pose. This may be due to the fact that extreme poses rarely appear in the dataset.
\\

\noindent{\bf Acknowledgement.}
This work was partly supported by the National Natural Science Foundation of China (Grant No. U1903213) and the Shenzhen Science and Technology Program (JCYJ20220818101014030).

{\small
\bibliographystyle{ieee_fullname}
\bibliography{main}

\begin{thebibliography}{10}\itemsep=-1pt

\bibitem{bansal2018recycle}
Aayush Bansal, Shugao Ma, Deva Ramanan, and Yaser Sheikh.
\newblock Recycle-gan: Unsupervised video retargeting.
\newblock In {\em Proceedings of the European conference on computer vision
  (ECCV)}, pages 119--135, 2018.

\bibitem{3dmm}
Volker Blanz and Thomas Vetter.
\newblock A morphable model for the synthesis of 3d faces.
\newblock In {\em ACM SIGGRAPH}, pages 187--194, 1999.

\bibitem{booth_3d_2016}
James Booth, Anastasios Roussos, Stefanos Zafeiriou, Allan Ponniah, and David
  Dunaway.
\newblock A 3d morphable model learnt from 10,000 faces.
\newblock In {\em 2016 {IEEE} Conference on Computer Vision and Pattern
  Recognition ({CVPR})}, pages 5543--5552.
\newblock {ISSN}: 1063-6919.

\bibitem{bounareli_finding_2022}
Stella Bounareli, Vasileios Argyriou, and Georgios Tzimiropoulos.
\newblock Finding directions in {GAN}'s latent space for neural face
  reenactment.

\bibitem{chen_puppeteergan_2020}
Zhuo Chen, Chaoyue Wang, Bo Yuan, and Dacheng Tao.
\newblock {PuppeteerGAN}: Arbitrary portrait animation with semantic-aware
  appearance transformation.
\newblock In {\em 2020 {IEEE}/{CVF} Conference on Computer Vision and Pattern
  Recognition ({CVPR})}, pages 13515--13524. {IEEE}.

\bibitem{deng2019arcface}
Jiankang Deng, Jia Guo, Niannan Xue, and Stefanos Zafeiriou.
\newblock Arcface: Additive angular margin loss for deep face recognition.
\newblock In {\em Proceedings of the IEEE/CVF conference on computer vision and
  pattern recognition}, pages 4690--4699, 2019.

\bibitem{doukas_headgan_2021}
Michail~Christos Doukas, Stefanos Zafeiriou, and Viktoriia Sharmanska.
\newblock {HeadGAN}: One-shot neural head synthesis and editing.

\bibitem{drobyshev_megaportraits_2022}
Nikita Drobyshev, Jenya Chelishev, Taras Khakhulin, Aleksei Ivakhnenko, Victor
  Lempitsky, and Egor Zakharov.
\newblock {MegaPortraits}: One-shot megapixel neural head avatars.

\bibitem{heusel2017gans}
Martin Heusel, Hubert Ramsauer, Thomas Unterthiner, Bernhard Nessler, and Sepp
  Hochreiter.
\newblock Gans trained by a two time-scale update rule converge to a local nash
  equilibrium.
\newblock {\em Advances in neural information processing systems}, 30, 2017.

\bibitem{hong_depth-aware_2022}
Fa-Ting Hong, Longhao Zhang, Li Shen, and Dan Xu.
\newblock Depth-aware generative adversarial network for talking head video
  generation.

\bibitem{karras_style-based_2019}
Tero Karras, Samuli Laine, and Timo Aila.
\newblock A style-based generator architecture for generative adversarial
  networks.

\bibitem{karras_analyzing_2020}
Tero Karras, Samuli Laine, Miika Aittala, Janne Hellsten, Jaakko Lehtinen, and
  Timo Aila.
\newblock Analyzing and improving the image quality of {StyleGAN}.

\bibitem{kim_animeceleb_2022}
Kangyeol Kim, Sunghyun Park, Jaeseong Lee, Sunghyo Chung, Junsoo Lee, and
  Jaegul Choo.
\newblock {AnimeCeleb}: Large-scale animation {CelebHeads} dataset for head
  reenactment.

\bibitem{lee2020maskgan}
Cheng-Han Lee, Ziwei Liu, Lingyun Wu, and Ping Luo.
\newblock Maskgan: Towards diverse and interactive facial image manipulation.
\newblock In {\em Proceedings of the IEEE/CVF Conference on Computer Vision and
  Pattern Recognition}, pages 5549--5558, 2020.

\bibitem{nagrani_voxceleb_2017}
Arsha Nagrani, Joon~Son Chung, and Andrew Zisserman.
\newblock {VoxCeleb}: a large-scale speaker identification dataset.
\newblock In {\em Interspeech 2017}, pages 2616--2620.

\bibitem{narvekar2011no}
Niranjan~D Narvekar and Lina~J Karam.
\newblock A no-reference image blur metric based on the cumulative probability
  of blur detection (cpbd).
\newblock {\em IEEE Transactions on Image Processing}, 20(9):2678--2683, 2011.

\bibitem{ren_pirenderer_2021}
Yurui Ren, Ge Li, Yuanqi Chen, Thomas~H. Li, and Shan Liu.
\newblock {PIRenderer}: Controllable portrait image generation via semantic
  neural rendering.
\newblock In {\em 2021 {IEEE}/{CVF} International Conference on Computer Vision
  ({ICCV})}, pages 13739--13748. {IEEE}.

\bibitem{siarohin_animating_2019}
Aliaksandr Siarohin, Stéphane Lathuilière, Sergey Tulyakov, Elisa Ricci, and
  Nicu Sebe.
\newblock Animating arbitrary objects via deep motion transfer.
\newblock version: 2.

\bibitem{siarohin_first_2020}
Aliaksandr Siarohin, Stéphane Lathuilière, Sergey Tulyakov, Elisa Ricci, and
  Nicu Sebe.
\newblock First order motion model for image animation.

\bibitem{song_everythings_2021}
Linsen Song, Wayne Wu, Chaoyou Fu, Chen Qian, Chen~Change Loy, and Ran He.
\newblock Everything's talkin': Pareidolia face reenactment.

\bibitem{tao_motion_2022}
Jiale Tao, Biao Wang, Tiezheng Ge, Yuning Jiang, Wen Li, and Lixin Duan.
\newblock Motion transformer for unsupervised image animation.

\bibitem{wang_one-shot_2021}
Ting-Chun Wang, Arun Mallya, and Ming-Yu Liu.
\newblock One-shot free-view neural talking-head synthesis for video
  conferencing.
\newblock In {\em 2021 {IEEE}/{CVF} Conference on Computer Vision and Pattern
  Recognition ({CVPR})}, pages 10034--10044. {IEEE}.

\bibitem{wang_latent_2022}
Yaohui Wang, Di Yang, Francois Bremond, and Antitza Dantcheva.
\newblock {LATENT} {IMAGE} {ANIMATOR}: {LEARNING} {TO} {ANIMATE} {IMAGES} {VIA}
  {LATENT} {SPACE} {NAVIGATION}.
\newblock page~17.

\bibitem{morph_animate}
{Wikipedia contributors}.
\newblock Morph target animation --- {Wikipedia}{,} the free encyclopedia.
\newblock
  \url{https://en.wikipedia.org/w/index.php?title=Morph_target_animation&oldid=1078665970},
  2022.
\newblock [Online; accessed 6-November-2022].

\bibitem{ferrari_x2face_2018}
Olivia Wiles, A.~Sophia Koepke, and Andrew Zisserman.
\newblock X2face: A network for controlling face generation using images,
  audio, and pose codes.
\newblock In Vittorio Ferrari, Martial Hebert, Cristian Sminchisescu, and Yair
  Weiss, editors, {\em Computer Vision – {ECCV} 2018}, volume 11217, pages
  690--706. Springer International Publishing.
\newblock Series Title: Lecture Notes in Computer Science.

\bibitem{wu_reenactgan_2018}
Wayne Wu, Yunxuan Zhang, Cheng Li, Chen Qian, and Chen~Change Loy.
\newblock {ReenactGAN}: Learning to reenact faces via boundary transfer.

\bibitem{xing2023codetalker}
Jinbo Xing, Menghan Xia, Yuechen Zhang, Xiaodong Cun, Jue Wang, and Tien-Tsin
  Wong.
\newblock Codetalker: Speech-driven 3d facial animation with discrete motion
  prior.
\newblock In {\em Proceedings of the IEEE/CVF Conference on Computer Vision and
  Pattern Recognition}, pages 12780--12790, 2023.

\bibitem{xu2022motion}
Borun Xu, Biao Wang, Jinhong Deng, Jiale Tao, Tiezheng Ge, Yuning Jiang, Wen
  Li, and Lixin Duan.
\newblock Motion and appearance adaptation for cross-domain motion transfer.
\newblock In {\em Computer Vision--ECCV 2022: 17th European Conference, Tel
  Aviv, Israel, October 23--27, 2022, Proceedings, Part XVI}, pages 529--545.
  Springer, 2022.

\bibitem{yin_styleheat_2022}
Fei Yin, Yong Zhang, Xiaodong Cun, Mingdeng Cao, Yanbo Fan, Xuan Wang, Qingyan
  Bai, Baoyuan Wu, Jue Wang, and Yujiu Yang.
\newblock {StyleHEAT}: One-shot high-resolution editable talking face
  generation via pre-trained {StyleGAN}.

\bibitem{zakharov_few-shot_2019}
Egor Zakharov, Aliaksandra Shysheya, Egor Burkov, and Victor Lempitsky.
\newblock Few-shot adversarial learning of realistic neural talking head
  models.
\newblock In {\em 2019 {IEEE}/{CVF} International Conference on Computer Vision
  ({ICCV})}, pages 9458--9467. {IEEE}.

\bibitem{zhang_freenet_2020}
Jiangning Zhang, Xianfang Zeng, Mengmeng Wang, Yusu Pan, Liang Liu, Yong Liu,
  Yu Ding, and Changjie Fan.
\newblock {FReeNet}: Multi-identity face reenactment.
\newblock In {\em 2020 {IEEE}/{CVF} Conference on Computer Vision and Pattern
  Recognition ({CVPR})}, pages 5325--5334. {IEEE}.

\bibitem{zhang2018unreasonable}
Richard Zhang, Phillip Isola, Alexei~A Efros, Eli Shechtman, and Oliver Wang.
\newblock The unreasonable effectiveness of deep features as a perceptual
  metric.
\newblock In {\em Proceedings of the IEEE conference on computer vision and
  pattern recognition}, pages 586--595, 2018.

\bibitem{zhang2023sadtalker}
Wenxuan Zhang, Xiaodong Cun, Xuan Wang, Yong Zhang, Xi Shen, Yu Guo, Ying Shan,
  and Fei Wang.
\newblock Sadtalker: Learning realistic 3d motion coefficients for stylized
  audio-driven single image talking face animation.
\newblock In {\em Proceedings of the IEEE/CVF Conference on Computer Vision and
  Pattern Recognition}, pages 8652--8661, 2023.

\end{thebibliography}
}


\clearpage
\onecolumn
\subsection*{A. Network Architecture Details}
\begin{figure*}
\centering
\includegraphics[width=\linewidth]{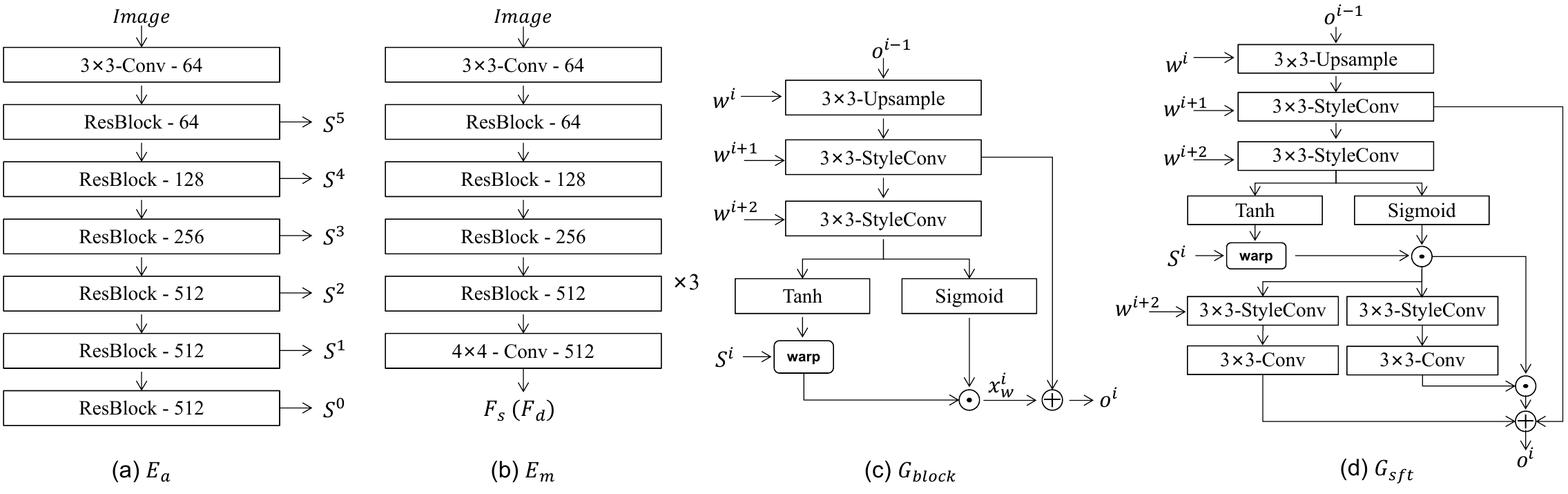}
\caption{The architecture of our ToonTalker. }
\label{fig:detail}
\end{figure*}

The architectures of several key model modules are shown in Fig.~\ref{fig:detail}.
Please note that the module structures of the real and cartoon domains are the same.
``ResBlock-64'' denotes the number of feature map channels in the residual block is $64$. 

\vspace{1mm}
\noindent\textbf{Appearance encoder $E_a$.} 
As shown in Fig.~\ref{fig:detail}(a), our appearance encoder consists of six ResBlocks. 
In each block the size of feature maps will be downsampled. 
Thus we can obtain feature maps with the different sizes, \textit{i.e.},  8$\times$8, 16$\times$16,  32$\times$32, 64$\times$64, 128$\times$128, and 256$\times$256. 

\vspace{1mm}
\noindent\textbf{Motion encoder $E_m$.} 
As shown in Fig.~\ref{fig:detail}(b), the motion encoder is composed of seven ResBlocks. 
And we use a 4$\times$4 convolution to downsample the feature map after the sever ResBlocks. 
The output motion code can be denoted as $F\in R^{512}$. 

\vspace{1mm}
\noindent\textbf{Motion Query Transformer $T$.} 
We use transformers to align motion spaces with common query tokens for their ability of catching long-range dependencies. In our model, as shown \textit{in Fig.3 of the manuscript}, there are two query transformers,  \textit{i.e.,} source query transformer and driving query transformer. The architectures of the two transformers are the same. The motion base consists of 20 learnable embeddings, denoted as $p = \{p^k\}^K_{k=1}, K=20$ and $p^k \in R^d, d=512$. 

\vspace{1mm}
\noindent\textbf{Backward Transformer $T_B$.} As shown \textit{in Fig.3 of the manuscript}, the backward transformer consists of three transformer blocks, which correspond to multi-scale feature maps.  

\vspace{1mm}
\noindent\textbf{Generator $G$.} 
StyleConv blocks are proved to be effective in StyleGAN. So similar to the architecture of StyleGAN, the generator contains six StyleConv blocks. 
In the first three blocks, the low-resolution feature map from $E_a$ will be warped and upsampled on the condition of the motion code. 
And in the last three blocks, we not only warp the feature map but also refine the warped feature by spatial feature transformation. 
The two types of modified StyleConv blocks are shown in Fig.~\ref{fig:detail}(c) and Fig.~\ref{fig:detail}(d).

\subsection*{B. More Qualitative Results.}
\begin{figure*}[t]
\centering
\includegraphics[width=\linewidth]{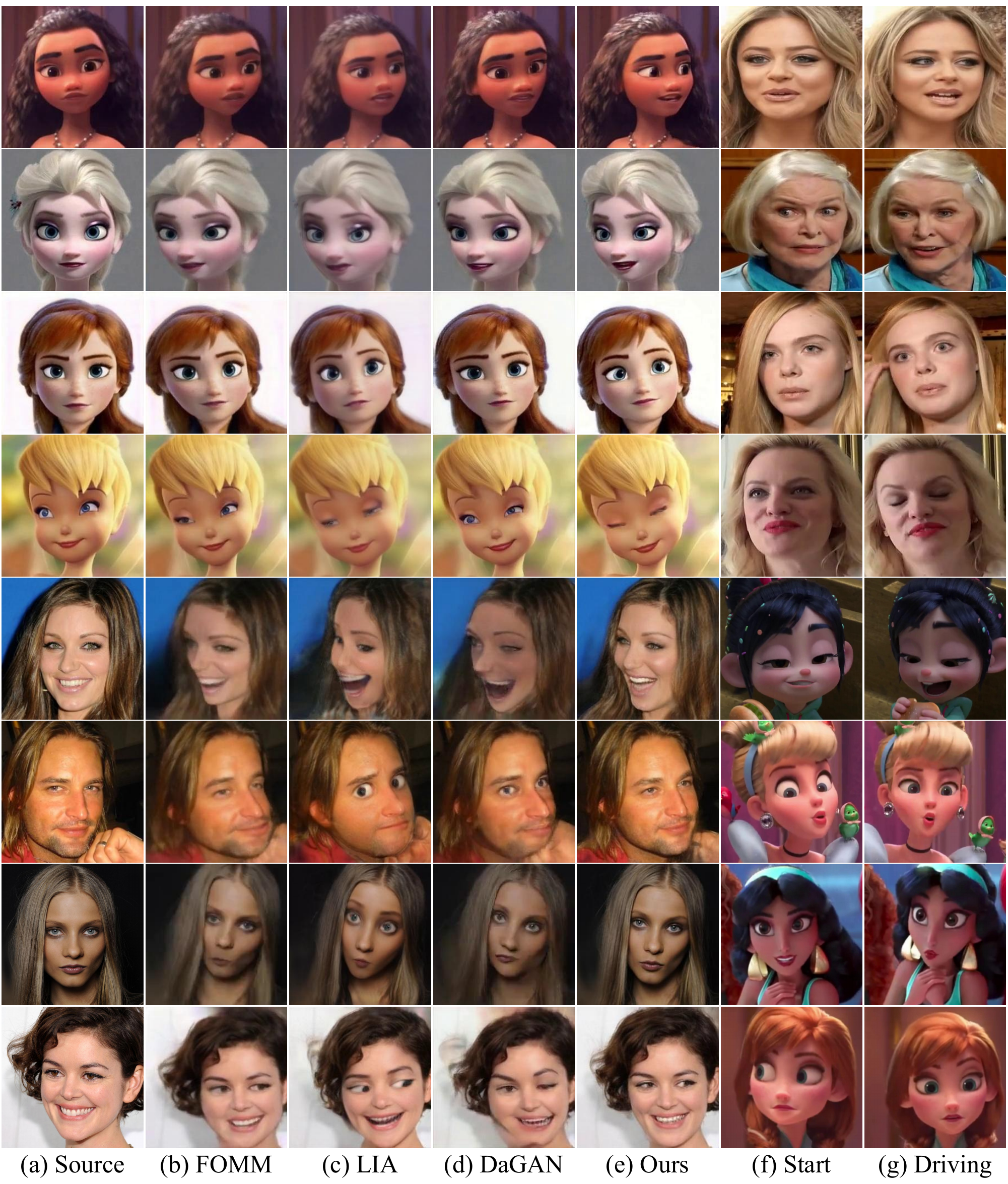}
\caption{Qualitative comparisons with \textit{state-of-the-art} methods on cross-domain reenactment. }
\label{fig:cross-domain-sup}
\end{figure*}

We show more examples of comparison with \textit{state-of-the-art} methods on cross-domain reenactment in Fig~\ref{fig:cross-domain-sup}.

\subsection*{C. Evaluation on Images Generated by Stable Diffusion }
\begin{figure*}[htb]
\centering
\includegraphics[width=0.9\linewidth]{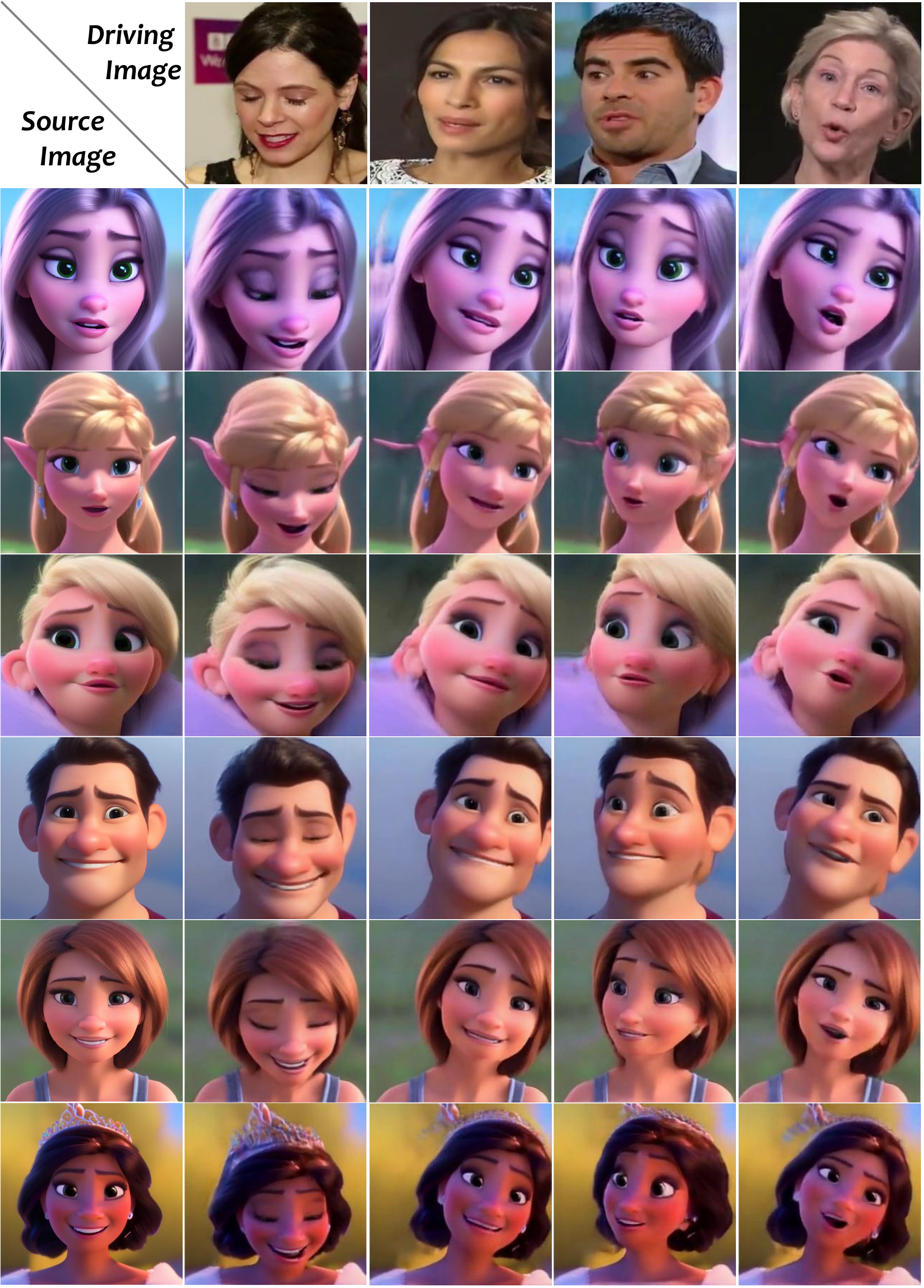}
\caption{Animating characters generated by Stable Diffusion.}
\label{fig:diffusion}
\end{figure*}

The fantastic generation ability of diffusion models has set off an upsurge in the world. 
Therefore, we use several cartoon characters generated by diffusion models as source images for cross-domain reenactment. We show some samples animated by our method in Fig.~\ref{fig:diffusion}. 




\end{document}